\documentclass[a4paper,12pt]{article}

\usepackage{comment}
\usepackage{fancyvrb}
\usepackage{graphicx}
\usepackage{hyperref}
\usepackage[ragged]{footmisc}
\usepackage{ragged2e}
\usepackage[section]{placeins}
\usepackage{caption}

\begin{document}

\title{\Huge INTRODUCTION TO THE SP THEORY OF INTELLIGENCE\\
{\Large Aiming to simplify and integrate observations and concepts across artificial intelligence, mainstream computing, mathematics, and human learning, perception, and cognition.}}

\author{J Gerard Wolff\footnote{Dr Gerry Wolff, BA (Cantab), PhD (Wales), CEng, MBCS, MIEEE; CognitionResearch.org, Menai Bridge, UK; \href{mailto:jgw@cognitionresearch.org}{jgw@cognitionresearch.org}; +44 (0) 1248 712962; +44 (0) 7746 290775; {\em Skype}: gerry.wolff; {\em Web}: \href{http://www.cognitionresearch.org}{www.cognitionresearch.org}.}}

\maketitle

\begin{abstract}

This article provides a brief introduction to the {\em SP Theory of Intelligence} and its realisation in the {\em SP Computer Model}. The overall goal of the SP programme of research, in accordance with long-established principles in science, has been the simplification and integration of observations and concepts across artificial intelligence, mainstream computing, mathematics, and human learning, perception, and cognition. In broad terms, the SP system is a brain-like system that takes in ``New'' information through its senses and stores some or all of it as ``Old'' information. A central idea in the system is the powerful concept of {\em SP-multiple-alignment}, borrowed and adapted from bioinformatics. This the key to the system's versatility in aspects of intelligence, in the representation of diverse kinds of knowledge, and in the seamless integration of diverse aspects of intelligence and diverse kinds of knowledge, in any combination. There are many potential benefits and applications of the SP system. It is envisaged that the system will be developed as the {\em SP Machine}, which will initially be a software virtual machine, hosted on a high-performance computer, a vehicle for further research and a step towards the development of an industrial-strength SP Machine. 

\end{abstract}

\section{Introduction}

The {\em SP theory of intelligence} and its realisation in the {\em SP computer model} is a system that has been under development since about 1987, with a break between early 2006 and late 2012.

Potential benefits and applications of the SP system include versatility in aspects of intelligence, versatility in the representation of diverse forms of knowledge, seamless integration of aspects of intelligence and kinds of knowledge in any combination, helping to solve nine problems with big data, providing a model for aspects of neuroscience, solving several problems with deep learning, and many more.

A key idea in the SP system is the powerful concept of {\em SP-multiple-alignment}, borrowed and adapted from the concept of `multiple sequence alignment' in bioinformatics. This may prove to be as significant for an understanding of human intelligence as is DNA in biological sciences: SP-multiple-alignment may prove to be the ``double helix'' of intelligence.

\begin{figure}[!htbp]
\captionsetup{format=plain}
\captionsetup{format=plain}
\centering
\includegraphics[width=0.9\textwidth]{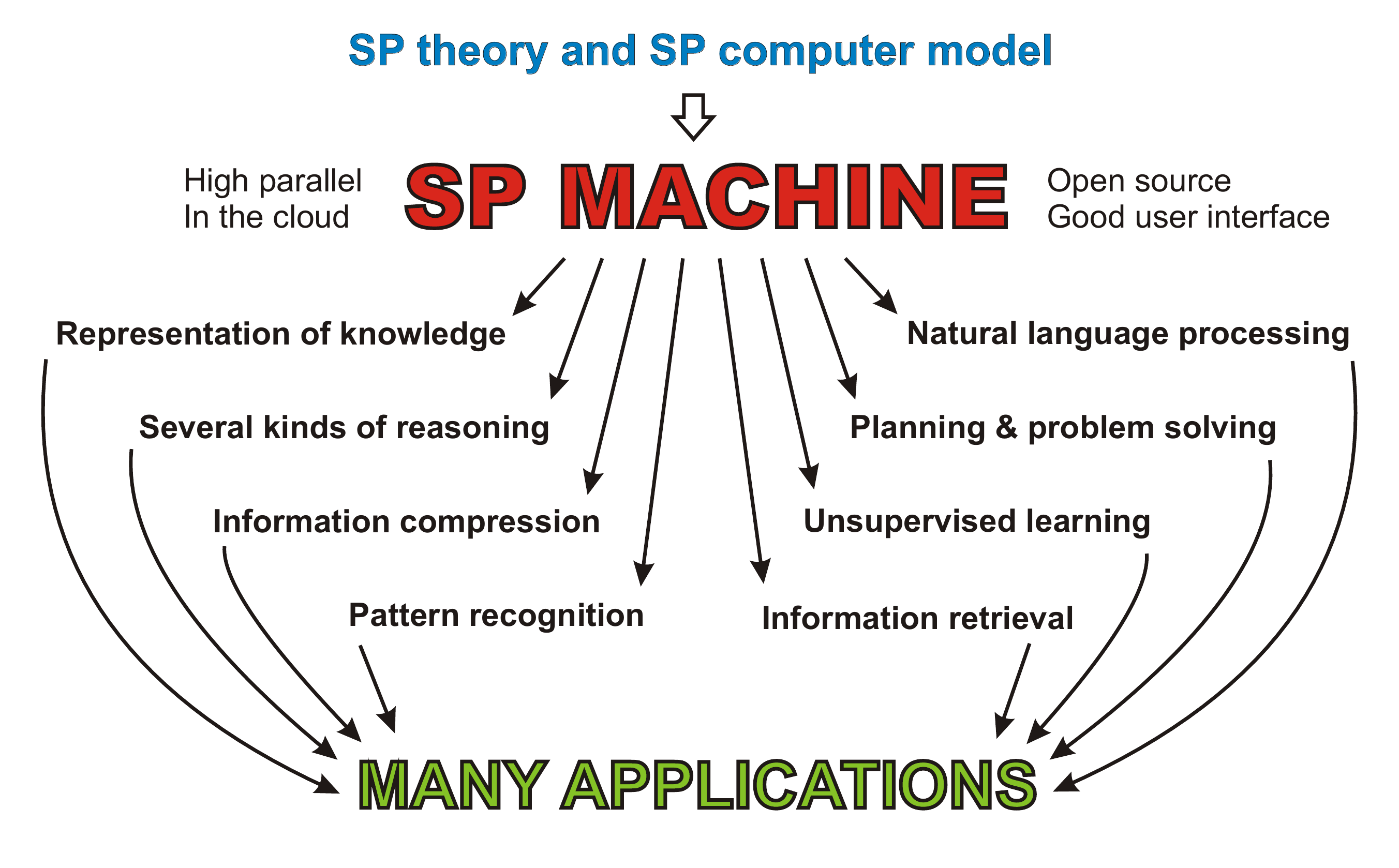}
\caption{Schematic representation of the development and application of the SP machine. Reproduced from Figure 2 in \cite{sp_extended_overview}, with permission.}
\label{sp_machine_figure}
\end{figure}

Although there are still residual problems to be solved \cite[Section 3.3]{sp_extended_overview}, it is envisaged that the SP computer model will be the basis for an {\em SP machine} as a software virtual machine, hosted on a high-performance computer. This would be a vehicle for further research and a basis for an industrial-strength SP machine with many potential benefits and application, as shown schematically in Figure \ref{sp_machine_figure}. A programme of development for the SP machine is described in \cite{devt_sp_machine}.

\section{Background}\label{background_section}

The overall goal of the SP programme of research, in accordance with long-established principles in science, has been the simplification and integration of observations and concepts across artificial intelligence, mainstream computing, mathematics, and human learning, perception, and cognition.

From the beginning, a unifying theme in the SP research has been that all kinds of processing would be done by compression of information via a search for patterns that match each other and via the merging or `unification' of patterns that are the same. The reason for the emphasis on the matching and unification of patterns is that this seems to provide a better handle on possible mechanisms in natural or artificial systems than do the more mathematically-oriented approaches to information compression.

The main motivation for the focus on information compression is research by Fred Attneave \cite{attneave_1954}, Horace Barlow \cite{barlow_1959,barlow_1969}, and others, showing the importance of information compression in the workings of brains and nervous systems. Solomonoff's seminal work on the development of {\em algorithmic probability theory} \cite{solomonoff_1964,solomonoff_1997} is also important.

Since people often ask, the name ``SP'' stands for {\em Simplicity} and {\em Power}, two ideas which, together, mean the same as information compression. This is because information compression may be seen to be a process of maximising `simplicity' in a body of information, by reducing redundancy in that information, whilst at the same time retaining as much as possible of its non-redundant expressive `power'.

\section{The SP system}\label{the_sp_system_section}

The SP system is described in outline here and in Appendix I of \cite{sp_alternatives}, in more detail in \cite{sp_extended_overview}, and in even more detail in \cite{wolff_2006}. Distinctive features and advantages of the SP system are described in \cite{sp_alternatives}.

\begin{figure}[!htbp]
\captionsetup{format=plain}
\centering
\includegraphics[width=0.5\textwidth]{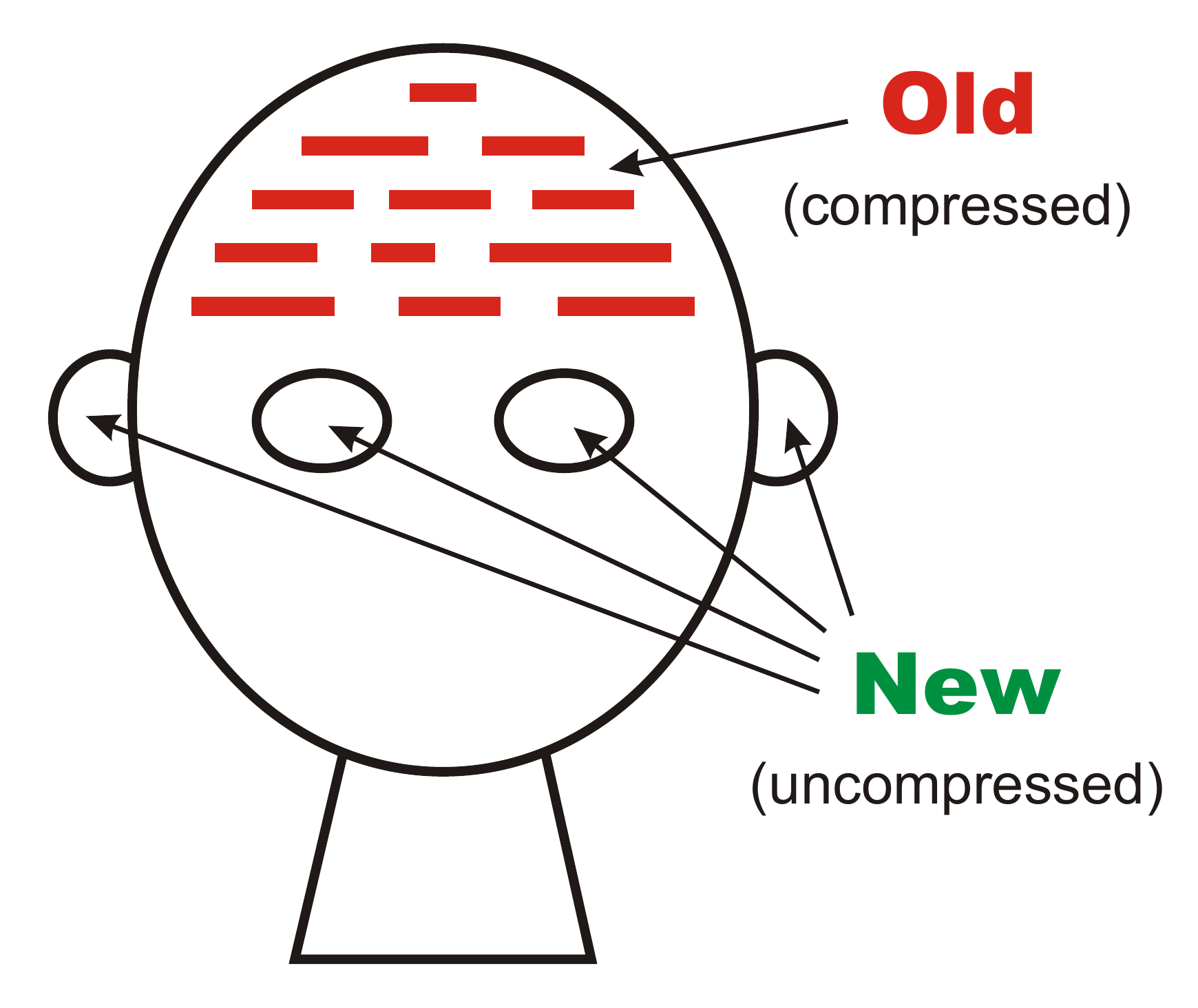}
\caption{Schematic representation of the SP system from an `input' perspective. Reproduced, with permission, from Figure 1 in \cite{sp_extended_overview}.}
\label{sp_input_perspective_figure}
\end{figure}

In broad terms, the SP system is a brain-like system that takes in {\em New} information through its senses and stores some or all of it as {\em Old} information, as shown schematically in Figure \ref{sp_input_perspective_figure}.

\subsection{SP-multiple-alignment}

A central idea in the SP system, is the concept of {\em SP-multiple-alignment}, borrowed and adapted from the concept of `multiple sequence alignment' in bioinformatics. Probably the best way to explain the idea is by way of examples, shown in Figures \ref{ma_dna_figure} and \ref{fortune_brave_multiple_alignment_figure}.

Figure \ref{ma_dna_figure} shows an example of multiple sequence alignment in bioinformatics. Here, there are five DNA sequences which have been arranged one above the other, and then, by judicious `stretching' of one or more of the sequences in a computer, symbols that match each other across two or more sequences have been brought into line. A `good' multiple sequence alignment, like the one shown, is one with a relatively large number of matching symbols from row to row. The process of discovering a good multiple sequence alignment is normally too complex to be done by exhaustive search, so heuristic methods are needed, building multiple sequence alignments in stages and, at each stage, selecting the best partial structures for further processing.

\begin{figure}[!htbp]
\captionsetup{format=plain}
\fontsize{10.00pt}{12.00pt}
\centering
{\bf
\begin{BVerbatim}
  G G A     G     C A G G G A G G A     T G     G   G G A
  | | |     |     | | | | | | | | |     | |     |   | | |
  G G | G   G C C C A G G G A G G A     | G G C G   G G A
  | | |     | | | | | | | | | | | |     | |     |   | | |
A | G A C T G C C C A G G G | G G | G C T G G A | G A
  | | |           | | | | | | | | |   |   |     |   | | |
  G G A A         | A G G G A G G A   | A G     G   G G A
  | |   |         | | | | | | | |     |   |     |   | | |
  G G C A         C A G G G A G G     C   G     G   G G A
\end{BVerbatim}
}
\caption{A `good' multiple alignment amongst five DNA sequences. Reproduced with permission from Figure 3.1 in \cite{wolff_2006}.}
\label{ma_dna_figure}
\end{figure}

Figure \ref{fortune_brave_multiple_alignment_figure} shows an example of an SP-multiple-alignment, superficially similar to the one in Figure \ref{ma_dna_figure}, except that sequences are called {\em SP-patterns} and, more importantly, one of the SP-patterns is New information and is normally shown in row 0, while the remaining SP-patterns are Old information, and these are shown in the remaining rows.

\begin{figure}[!htbp]
\captionsetup{format=plain}
\fontsize{06.00pt}{07.20pt}
\centering
{\bf
\begin{BVerbatim}
0              f o r t u n e                      f a v o u r     s             t h e        b r a v e               0
               | | | | | | |                      | | | | | |     |             | | |        | | | | |
1              | | | | | | |                 Vr 6 f a v o u r #Vr |             | | |        | | | | |               1
               | | | | | | |                 |                 |  |             | | |        | | | | |
2              | | | | | | |             V 7 Vr               #Vr s #V          | | |        | | | | |               2
               | | | | | | |             |                          |           | | |        | | | | |
3              | | | | | | |        VP 3 V                          #V NP       | | |        | | | | |    #NP #VP    3
               | | | | | | |        |                                  |        | | |        | | | | |     |   |
4          N 4 f o r t u n e #N     |                                  |        | | |        | | | | |     |   |     4
           |                 |      |                                  |        | | |        | | | | |     |   |
5     NP 2 N                 #N #NP |                                  |        | | |        | | | | |     |   |     5
      |                          |  |                                  |        | | |        | | | | |     |   |
6 S 0 NP                        #NP VP                                 |        | | |        | | | | |     |  #VP #S 6
                                                                       |        | | |        | | | | |     |
7                                                                      |        | | |    N 5 b r a v e #N  |         7
                                                                       |        | | |    |             |   |
8                                                                      NP 1 D   | | | #D N             #N #NP        8
                                                                            |   | | | |
9                                                                           D 8 t h e #D                             9
\end{BVerbatim}
}
\caption{The best SP-multiple-alignment produced by the SP computer model with a New SP-pattern, `\texttt{f o r t u n e f a v o u r s t h e b r a v e}', representing a sentence to be parsed and a repository of user-supplied Old SP-patterns representing grammatical categories, including morphemes and words. Reproduced with permission from Figure 2 in \protect\cite{spneural_2016}.}
\label{fortune_brave_multiple_alignment_figure}
\end{figure}

As can be seen from this example, the building of an SP-multiple-alignment may achieve the effect of parsing a sentence (`\texttt{f o r t u n e f a v o u r s t h e b r a v e}' in this example) into its grammatical parts and sub-parts. But as we shall see later, the SP system has strengths in several different aspects of intelligence, and in the representation of several different kinds of knowledge---and most of this versatility flows from the building of SP-multiple-alignments.

To create an SP-multiple-alignment like the one shown in Figure \ref{fortune_brave_multiple_alignment_figure}, the SP system starts with a relatively large repository of Old SP-patterns, each one representing a syntactic structure in English, which may be a morpheme, a word, or a higher-level structure. The Old SP-patterns would ideally be learned by the system, but pending full development of the learning processes, the Old SP-patterns may be supplied to the system by the user.

With a repository of Old SP-patterns in place, the SP system is supplied with the New SP-pattern (`\texttt{f o r t u n e f a v o u r s t h e b r a v e}') and the system tries to build one or more SP-multiple-alignments, each of which allows the New SP-pattern to be encoded economically in terms the Old SP-patterns in the SP-multiple-alignment.

The details of how the encoding is done need not detain us here, but it is relevant to note that the SP-multiple-alignment construct, in conjunction with unsupervised learning in the SP system (outlined below), appears to provide a means of achieving relatively high levels of information compression with many kinds of data.

As with the building of `good' multiple sequence alignments in bioinformatics, the creation of one or more `good' SP-multiple-alignments is normally too complex to be done by any exhaustive process. As with multiple sequence alignments in bioinformatics, heuristic search is needed, building SP-multiple-alignments in stages and, at each stage, selecting the best partial structures for further processing. With this approach, it is not normally possible to guarantee that the best possible SP-multiple-alignment has been found, but it is normally possible to create SP-multiple-alignments that are `good enough'.

\subsection{Learning in the SP system}\label{learning_section}

In the SP system, learning is special. Instead of it being a by-product of the building of SP-multiple-alignments it is a process of creating {\em grammars}, where each grammar is a collection of Old SP-patterns (many of which would normally be derived from partial matches between SP-patterns within SP-multiple-alignments), and each grammar is scored in terms of its effectiveness via SP-multiple-alignment in the economical encoding of a target set of New SP-patterns. As with the building of SP-multiple-alignments, the process is too complex for exhaustive search so heuristic methods are needed.

In the SP system, learning is normally `unsupervised', deriving structures from incoming sensory information without the need for any kind of `teacher', or `reinforcement', or anything equivalent. But in case this seems unduly narrow, it appears that unsupervised learning is the most general kind of learning and that, within the framework of unsupervised learning in the SP system, there is potential to model other kinds of learning such as `supervised' learning, `reinforcement' learning, and more.

\subsection{SP-neural}\label{sp-neural_section}

\begin{figure}[!htbp]
\captionsetup{format=plain}
\centering
\includegraphics[width=0.7\textwidth]{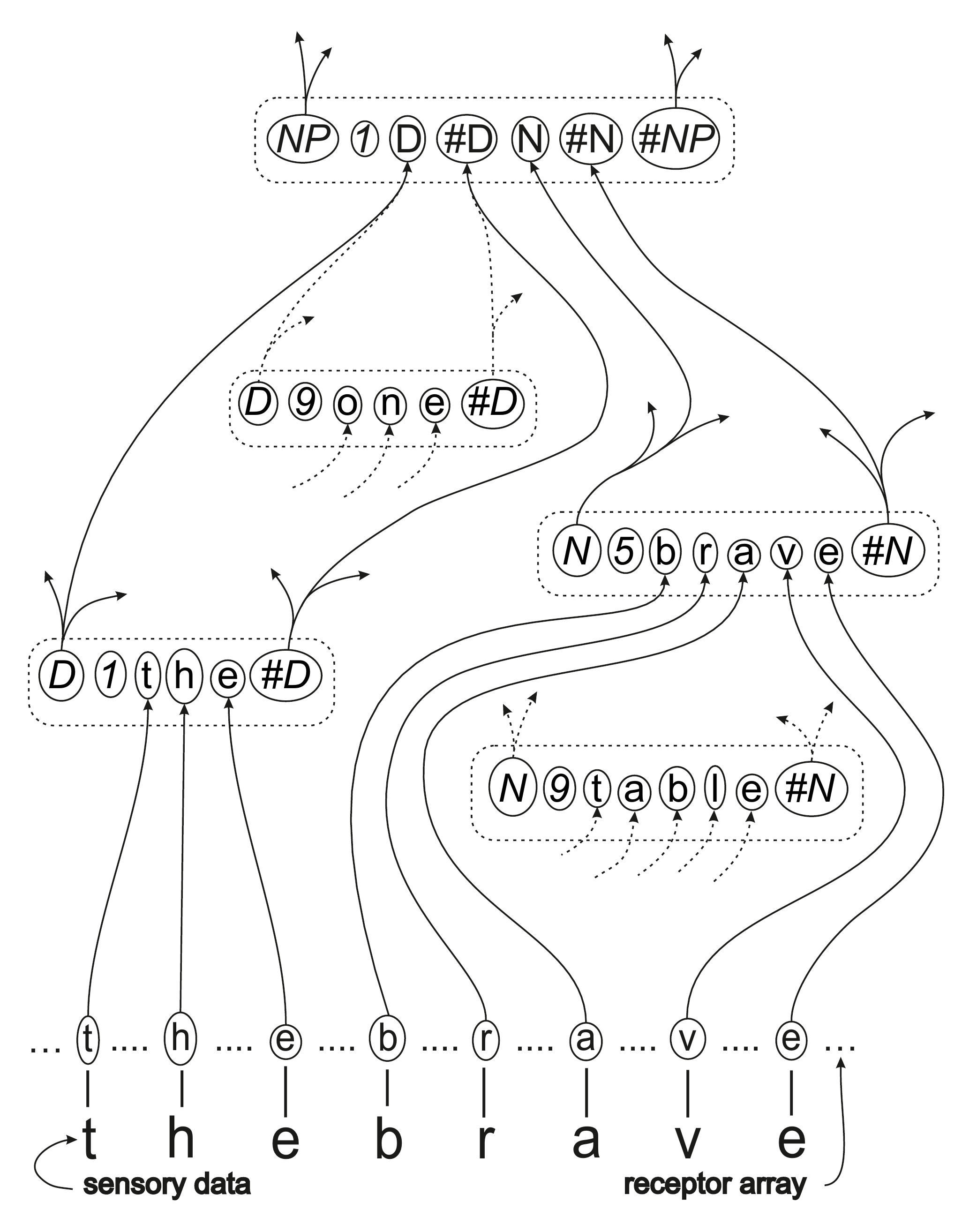}
\caption{A schematic representation of a partial SP-multiple-alignment in SP-neural, as discussed in \cite[Section 4]{spneural_2016}. Each broken-line rectangle with rounded corners represents a {\em pattern assembly}---corresponding to an SP-pattern in the main SP theory; each character or group of characters enclosed in a solid-line ellipse represents a {\em neural symbol} corresponding to an SP-symbol in the main SP theory; the lines between pattern assemblies represent nerve fibres with arrows showing the direction in which impulses travel; neural symbols are mainly symbols from linguistics such as `\texttt{NP}' meaning `noun phrase, `\texttt{D}' meaning a `determiner', `\texttt{\#D}' meaning the end of a determiner, `\texttt{\#NP}' meaning the end of a noun phrase, and so on. Reproduced with permission from Figure 3 in \cite{spneural_2016}.}
\label{the_brave_neural_figure}
\end{figure}

A potentially useful feature of the SP system is that it is possible to see how abstract constructs and processes in the system may be realised in terms of neurons and their interconnections. This is the basis for {\em SP-neural}, a `neural' version of the SP system, described in an early form in \cite[Chapter 11]{wolff_2006}, and in an updated and more detailed form in \cite{spneural_2016}, and illustrated in Figure \ref{the_brave_neural_figure}.

In this connection, it is relevant to mention that the SP system, in both its abstract and neural forms, is quite different from deep learning in artificial neural networks \cite{schmidhuber_2015} and has substantial advantages compared with such systems, as described in \cite[Section V]{sp_alternatives} and in \cite{spdlsol_2018}. Some examples of those advantages are outlined in Section \ref{other_strengths_section}, below.

\subsection{Generalising the SP system for two-dimensional SP-patterns, both static and moving}

This brief description of the SP system and how it works may have given the impression that it is intended to work entirely with sequences of SP-symbols, like multiple sequence alignments in bioinformatics. But it is envisaged that, in future development of the system, two-dimensional SP-patterns will be introduced, with potential to represent and process such things as photographs and diagrams, and structures in three dimensions as described in \cite[Section 6.1 and 6,2]{sp_vision}, and procedures that work in parallel as described in \cite[Sections V-G, V-H, and V-I, and Appendix C]{sp_autonomous_robots}.

It is envisaged that, at some stage, the SP system will be generalised to work with sequences of two-dimensional `frames' from moving visual media.

\section{Versatility in aspects of intelligence}\label{versatility_in_aspects of_intelligence_section}

Strengths and potential of the SP system are summarised in this section and in those that follow. Further information may be found in \cite[Sections 5 to 12]{sp_extended_overview}, \cite[Chapters 5 to 9]{wolff_2006}, \cite{sp_alternatives}, and in other sources referenced in the sections that follow. This section outlines the SP system's versatility in aspects of intelligence.

The system has, first, strengths and potential in the `unsupervised' learning of new knowledge. As we saw in Section \ref{learning_section}, this is an aspect of intelligence in the SP system that is different from others because it is not a by-product of the building of multiple alignments but is, instead, achieved via the creation of {\em grammars}, drawing on information within SP-multiple-alignments.

Secondly, other aspects of intelligence exhibited by the SP system are modelled via the building of SP-multiple-alignments. These other aspects of intelligence include: the analysis and production of natural language; pattern recognition that is robust in the face of errors in data; pattern recognition at multiple levels of abstraction; computer vision \cite{sp_vision}; best-match and semantic kinds of information retrieval; several kinds of reasoning (next paragraph); planning; and problem solving.

Thirdly, kinds of reasoning that may be modelled in the SP system include: one-step `deductive' reasoning; chains of reasoning; abductive reasoning; reasoning with probabilistic networks and trees; reasoning with `rules'; nonmonotonic reasoning and reasoning with default values; Bayesian reasoning with `explaining away'; causal reasoning; reasoning that is not supported by evidence; the inheritance of attributes in class hierarchies; and inheritance of contexts in part-whole hierarchies. Where it is appropriate, probabilities for inferences may be calculated in a straightforward manner (\cite[Section 3.7]{wolff_2006}, \cite[Section 4.4]{sp_extended_overview}).

There is also potential in the system for spatial reasoning \cite[Section IV-F.1]{sp_autonomous_robots}, and for what-if reasoning \cite[Section IV-F.2]{sp_autonomous_robots}.

It seems unlikely that the features of intelligence mentioned above are the full extent of the SP system's potential to imitate what people can do. The close connection that is known to exist between information compression and concepts of prediction and probability \cite{solomonoff_1964,solomonoff_1997,li_vitanyi_2014}, the central role of information compression in the SP-multiple-alignment framework, and the versatility of the SP-multiple-alignment framework in aspects of intelligence suggests that there there are more insights to come.

\section{Versatility in the representation of knowledge}\label{versatility_in_representation_of_knowledge_section}

Although SP-patterns are not very expressive in themselves, they come to life in the SP-multiple-alignment framework. Within that framework, they may serve in the representation of several different kinds of knowledge, including: the syntax of natural languages; class-inclusion hierarchies (with or without cross classification); part-whole hierarchies; discrimination networks and trees; if-then rules; entity-relationship structures \cite[Sections 3 and 4]{wolff_sp_intelligent_database}; relational tuples ({\em ibid}., Section 3), and concepts in mathematics, logic, and computing, such as `function', `variable', `value', `set', and `type definition' (\cite[Chapter 10]{wolff_2006}, \cite[Section 6.6.1]{sp_benefits_apps}, \cite[Section 2]{sp_software_engineering}).

As previously noted, the addition of two-dimensional SP patterns to the SP computer model is likely to expand the representational repertoire of the SP system to structures in two-dimensions and three-dimensions, and the representation of procedural knowledge with parallel processing.

As with the SP system's generality in aspects of intelligence, it seems likely that the SP system is not constrained to represent only the forms of knowledge that have been mentioned. The generality of information compression as a means of representing knowledge in a succinct manner, the central role of information compression in the SP-multiple-alignment framework, and the versatility of that framework in the representation of knowledge, suggest that the SP system may prove to be a means of representing {\em all} the kinds of knowledge that people may work with.

\section{Seamless integration of diverse aspects of intelligence, and diverse kinds of knowledge, in any combination}\label{seamless_integration_section}

An important third feature of the SP system, alongside its versatility in aspects of intelligence and its versatility in the representation of diverse kinds of knowledge, is that {\em there is clear potential for the SP system to provide seamless integration of diverse aspects of intelligence and diverse kinds of knowledge, in any combination.} This is because diverse aspects of intelligence and diverse kinds of knowledge all flow from a single coherent and relatively simple source: the SP-multiple-alignment framework.

It appears that seamless integration of diverse aspects of intelligence and diverse kinds of knowledge, in any combination, is {\em essential} in any artificial system that aspires to the fluidity, versatility and adaptability of the human mind.

\begin{figure}[!hbt]
\captionsetup{format=plain}
\centering
\includegraphics[width=0.8\textwidth]{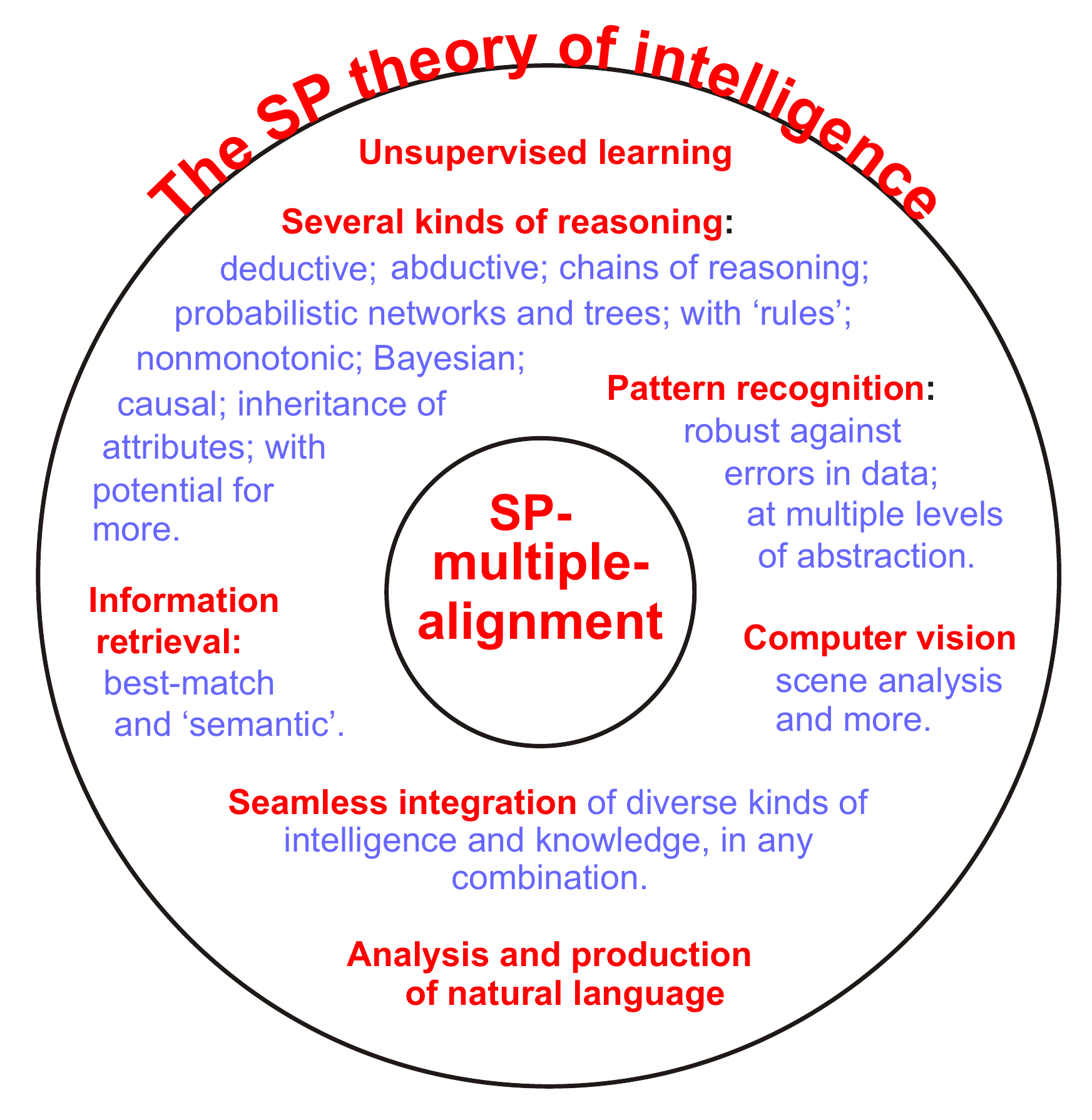}
\caption{A schematic representation of versatility and integration in the SP system, with SP-multiple-alignment centre stage.}
\label{versatility_integration_figure}
\end{figure}

Figure \ref{versatility_integration_figure} shows schematically how the SP system, with SP-multiple-alignment centre stage, exhibits versatility and integration.

\section{Potential benefits and applications of the SP system}\label{benefits_and_applications_section}

Apart from its strengths and potential in modelling aspects of the human mind, it appears that, in more humdrum terms, the SP system has several potential benefits and applications. These include:

\begin{itemize}

    \item {\em Big data}. Somewhat unexpectedly, it has been discovered that the SP system has considerable potential to help solve nine significant problems associated with big data \cite{sp_big_data}. These are: overcoming the problem of variety in big data; the unsupervised learning of structures and relationships in big data; interpretation of big data via pattern recognition, natural language processing and more; the analysis of streaming data; compression of big data; model-based coding for efficient transmission of big data; potential gains in computational and energy efficiency in the analysis of big data; managing errors and uncertainties in data; and visualisation of structure in big data and providing an audit trail in the processing of big data.

    \item {\em Autonomous robots}. The SP system opens up a radically new approach to the development of intelligence in autonomous robots \cite{sp_autonomous_robots};

    \item {\em An intelligent database system}. The SP system has potential in the development of an intelligent database system with several advantages compared with traditional database systems \cite{wolff_sp_intelligent_database}. In this connection, the SP system has potential to add several kinds of reasoning and other aspects of intelligence to the `database' represented by the World Wide Web, especially if the SP machine were to be supercharged by replacing the search mechanisms in the foundations of the SP machine with the high-parallel search mechanisms of any of the leading search engines.

    \item {\em Medical diagnosis}. The SP system may serve as a vehicle for medical knowledge and to assist practitioners in medical diagnosis, with potential for the automatic or semi-automatic learning of new knowledge \cite{wolff_medical_diagnosis};

    \item {\em Computer vision and natural vision}. The SP system opens up a new approach to the development of computer vision and its integration with other aspects of intelligence. It also throws light on several aspects of natural vision \cite{sp_vision};

    \item {\em Neuroscience}. As outlined in Section \ref{sp-neural_section}, abstract concepts in the SP theory map quite well into concepts expressed in terms of neurons and their interconnections in a version of the theory called {\em SP-neural} (\cite{spneural_2016}, \cite[Chapter 11]{wolff_2006}). This has potential to illuminate aspects of neuroscience and to suggest new avenues for investigation.

    \item \sloppy {\em Commonsense reasoning}. In addition to the previously-described strengths of the SP system in several kinds of reasoning, the SP system has strengths in the surprisingly challenging area of ``commonsense reasoning'', as described by Ernest Davis and Gary Marcus \cite{davis_marcus_2015}. How the SP system may meet the several challenges in this area is described in \cite{sp_csrk}.

    \item {\em Other areas of application}. The SP system has potential in several other areas of application including \cite{sp_benefits_apps}: the simplification and integration of computing systems; applications of natural language processing; best-match and semantic forms of information retrieval; software engineering \cite{sp_software_engineering}; the representation of knowledge, reasoning, and the semantic web; information compression; bioinformatics; the detection of computer viruses; and data fusion.

    \item {\em Mathematics}. The concept of information compression via the matching and unification of patterns provides an entirely novel interpretation of mathematics \cite{sp_maths_mystery}. This interpretation is quite unlike anything described in existing writings about the philosophy of mathematics or its application in science. There are potential benefits in science from this new interpretation of mathematics.

\end{itemize}

\section{Other strengths of the SP system}\label{other_strengths_section}

Apart the strengths of the SP system described in the previous four sections, there are others described mainly in \cite{sp_alternatives}. A selection of what appear to be the more important ones are outlined here:

\begin{itemize}

    \item {\em The amounts of data and processing required for learning}. There is now widespread recognition of the unreasonably large amounts of data that are currently required by deep learning systems to learn anything useful, and the correspondingly large amount of processing that is needed.\footnote{See, for example, ``Greedy, brittle, opaque, and shallow: the downsides to deep learning'', {\em Wired}, 2018-02-03, \href{http://bit.ly/2nM1ccg}{bit.ly/2nM1ccg}.} Although there is still work to be done in the development of unsupervised learning in the SP system, it is clear from the overall approach and from what has been achieved already that the learning of meaningful knowledge by the SP system is likely to require substantially less data and processing than with deep learning systems \cite[Sections V-D and V-E]{sp_alternatives}.

    \item {\em Transparency in the representation of knowledge, and in processing}. It is now widely recognised that a major problem with deep learning systems is that the way in which learned knowledge is represented in deep learning systems is far from being transparent and comprehensible by people, and that the way in which deep learning systems arrive at their conclusions is difficult or impossible for people to understand. These deficiencies are of concern for reasons of safety, legal liability, and more.\footnote{See, for example, ``Inside DARPA's push to make artificial intelligence explain itself'', {\em CET US News}, 2017-08-10, \href{http://bit.ly/2FQMoAr}{bit.ly/2FQMoAr}.}

        By contrast, and as with big data (Section \ref{benefits_and_applications_section}), knowledge in the SP system is represented in a manner that is familiar to people, using such devices as class-inclusion hierarchies, part-whole hierarchies, and others. And there is an audit trail for all processing in the SP system, so that it is explicit and comprehensible by people.

    \item {\em Catastrophic forgetting and continuous learning}. A major problem with deep learning---`catastrophic forgetting'---is that new leaning wipes out old learning. A related problem is that, to be practical, a learning system, like a person, should be able to learn continuously from its environment without old knowledge being disturbed by new knowledge.\footnote{Even if, for example, a person changes their name, the system should be able to retain both names and the date when the name was changed.}\textsuperscript{,}\footnote{It appears that this problem is a matter of concern to military planners as described, for example, in ``DARPA seeking AI that learns all the time'', {\em IEEE Spectrum}, 2017-11-21, \href{http://bit.ly/2BdERfZ}{bit.ly/2BdERfZ}.}

        A robust solution to this problem is intrinsic to the design of the SP system: new learning does not disturb old learning.

    \item {\em Learning with generalisation, and learning from `dirty data'}. A general problem with any system for unsupervised learning is how to generalise beyond the finite body of information that has been seen since the `birth' of the learning system, and how to correct over-generalisations and under-generalisations without the provision of a `teacher' or anything equivalent. A related problem is how `correct' knowledge may be learned from `dirty data', meaning data that contains `errors'.

        With systems for deep learning, several different solutions to the generalisation problem have been proposed, but none of them have any good theoretical underpinning \cite[Section V-H]{sp_alternatives}. It appears that, with deep learning systems, no solutions have been offered to the problem of learning from `dirty data'.

        By contrast, the SP system offers a relatively simple solution to both problems that derives directly from the unifying principe---compression of information---that lies at the heart of the SP system. How compression of information may generalise correctly from its raw data, and how it may achieve learning from `dirty data' is described in \cite[Section 9.5.3]{wolff_2006} and \cite[Section 5.3]{sp_extended_overview}. There is also relevant discussion in \cite[Sections V-H and XI-C]{sp_alternatives}. Experiments with computational models of learning suggest that the analysis is sound.

\end{itemize}

\section{Conclusion}

\sloppy It seems that the overarching goal of this research---the simplification and integration of observations and concepts across artificial intelligence, mainstream computing, mathematics, and human learning, perception, and cognition---has, to a large extent, been achieved.

The SP system provides a favourable combination of simplicity and power: the concept of SP-multiple-alignment, together with some relatively simple procedures for unsupervised learning, have proved to be remarkably versatile across diverse aspects of intelligence, in the representation of diverse kinds of knowledge, and in the seamless integration of diverse aspects of intelligence, and diverse kinds of knowledge, in any combination.

That last feature---seamless integration of diverse aspects of intelligence and diverse kinds of knowledge---appears to be essential in any artificial system that aspires to the fluidity, versatility, and adaptability of the human mind.

The SP system, compared with AI-related alternatives considered in \cite{sp_alternatives}, appears to provide a relatively firm foundation for the development of general human-like intelligence.

\bibliographystyle{plain}

\end{document}